\documentclass{article}

\usepackage{arxiv}
\usepackage{float} 
\usepackage[utf8]{inputenc} 
\usepackage[T1]{fontenc}    
\usepackage{url}            
\usepackage{booktabs}       
\usepackage{amsfonts}       
\usepackage{nicefrac}       
\usepackage{microtype}      
\usepackage{lipsum}		
\usepackage{graphicx}
\usepackage{natbib}
\usepackage{doi}
\usepackage{tabularx} 
\usepackage{booktabs} 
\usepackage{algorithm}
\usepackage{algpseudocode}
\usepackage{amsmath}
\usepackage{placeins}
\usepackage{makecell} 
\usepackage{authblk}  

\title{Modernizing CNN-based Weather Forecast Model towards Higher Computational Efficiency}

\author{
  Minjong Cheon\textsuperscript{1}, 
  Eunhan Goo\textsuperscript{2}, 
  Su-Hyeon Shin\textsuperscript{2}, 
  Muhammad Ahmed\textsuperscript{3}, 
  Hyungjun Kim\textsuperscript{*}\thanks{Corresponding author. Email: \texttt{hyungjun.kim@kaist.ac.kr}}
}

\date{
  \textsuperscript{1}Applied Science Research Institute, Korea Advanced Institute of Science and Technology (KAIST), Daejeon, South Korea\\
  \textsuperscript{2}Graduate School of Green Growth and Sustainability, KAIST, Daejeon, South Korea\\
  \textsuperscript{3}Department of Civil and Environmental Engineering, KAIST, Daejeon, South Korea\\
  \textsuperscript{*}Moon Soul Graduate School of Future Strategy, KAIST, Daejeon, South Korea\\[2ex]
  \today
}

\hypersetup{
pdftitle={A template for the arxiv style},
pdfsubject={q-bio.NC, q-bio.QM},
pdfauthor={Minjong},
pdfkeywords={First keyword, Second keyword, More},
}

\begin{document}
\maketitle

\begin{abstract}
Recently, AI-based weather forecast models have achieved impressive advances. These models have reached accuracy levels comparable to traditional NWP systems, marking a significant milestone in data-driven weather prediction. However, they mostly leverage Transformer-based architectures, which often leads to high training complexity and resource demands due to the massive parameter sizes. In this study, we introduce a modernized CNN-based model for global weather forecasting that delivers competitive accuracy while significantly reducing computational requirements. To present a systematic modernization roadmap, we highlight key architectural enhancements across multiple design scales from an earlier CNN-based approach. KAI-$\alpha$ incorporates a scale-invariant architecture and InceptionNeXt-based blocks within a geophysically-aware design, tailored to the structure of Earth system data. Trained on the ERA5 daily dataset with 67 atmospheric variables, the model contains about 7 million parameters and completes training in just 12 hours on a single NVIDIA L40s GPU. Our evaluation shows that KAI-$\alpha$ matches the performance of state-of-the-art models in medium-range weather forecasting, while offering a significantly lightweight design. Furthermore, case studies on the 2018 European heatwave and the East Asian summer monsoon demonstrate KAI-$\alpha$s robust skill in capturing extreme events, reinforcing its practical utility.

\end{abstract}

\section{Introduction}
\label{sec:intro}

Advancements in artificial intelligence (AI) have significantly reshaped the landscape of global weather forecasting. Traditional numerical weather prediction (NWP) models rely on solving complex atmospheric physics equations, demanding substantial computational resources and often requiring hours of runtime on supercomputers to generate forecasts\cite{pathak2022fourcastnet}. In contrast to traditional NWP models, emerging AI-driven forecasting systems utilize historical observational and simulation data to generate global forecasts rapidly, typically within seconds or minutes \cite{vaughan2024aardvark}. This not only significantly relieves computational burden and predictive responsiveness but also allows deep learning models to continually enhance their performance as they incorporate additional historical datasets.

Recent state-of-the-art AI weather models, such as Google DeepMind’s GraphCast, Huawei’s Pangu-Weather, NVIDIA’s FourCastNet, Microsoft’s Aurora, Fudan University’s FuXi, and Shanghai Artificial Intelligence Laboratory’s FengWu, exemplify this transformative shift in weather forecasting\cite{bi2023accurate}\cite{lam2022graphcast}\cite{chen2023fuxi}\cite{bodnar2024aurora}\cite{chen2023fengwu}\cite{pathak2022fourcastnet}. These models utilize diverse deep learning architectures such as Graph Neural Networks (GNN), Adaptive Fourier Neural Operator (AFNO), and Swin-Transformer. However, deep learning–based weather forecasting models utilize a very large number of parameters—typically ranging from tens of millions to over a billion\cite{pathak2022fourcastnet} \cite{bodnar2024aurora}\cite{chen2023fuxi}. These extensive parameter counts lead to high computational demands, making these models resource-intensive to run, particularly during training. Despite the computational cost, attention-based architectures have dominated recent advancements in AI weather forecasting due to their ability to capture long-range dependencies and spatiotemporal interactions \cite{ji2024spatio}. Nevertheless, comparatively fewer approaches have explored the use of Convolutional Neural Networks (CNNs) in this domain, even though CNN-based models are generally lighter and more efficient in terms of memory usage and inference speed \cite{nauen2023transformer}. This gap underscores the need to revisit CNN-based approaches that can offer a more efficient and scalable alternative to transformer-based architectures for deep learning based weather forecasting.

Therefore, we introduce the Korea Advanced Institute of Science and Technology (KAIST) AI model for Atmosphere (KAI-$\alpha$) model, an ultralight CNN-based model designed for global daily weather forecasting, requiring only around 7 million parameters, and just 12 hours of training on a single NVIDIA L40s GPU. The KAI-$\alpha$ architecture is inspired by recent advancements in lightweight convolutional backbones with scale-invariant structure and incorporates a design optimized for geospatial continuity, replacing zero padding with geocyclic padding to ensure seamless transitions across latitudinal and longitudinal boundaries \cite{cheon2024karina}. Trained on a daily 2.5-degree ERA5 dataset comprising 67 variables, KAI-$\alpha$ achieves competitive model performance compared to state-of-the-art AI-based implementations, despite its significantly reduced parameter count. Furthermore, building upon the foundational work of Weyn et al. (2020), we modernize the architecture and propose a roadmap toward a more computationally efficient paradigm for global weather forecasting \cite{weyn2020improving}. Specifically, our contributions are:

\begin{itemize}
    \setlength\itemsep{0.5em} 

    \item We introduce KAI-$\alpha$, built on a refined variant of the InceptionNeXt architecture, matching other state-of-the-art (SOTA) models in daily forecasting.
    
    \item We propose a roadmap for modernizing the CNN-based weather forecasting model.
    
    \item  KAI-$\alpha$ significantly reduces the number of parameters and requires only approximately 12 hours on a single NVIDIA L40S GPU.

\end{itemize}

\section{Methodology}

\subsection{Data Description}
Following the variables listed in Table 1, we utilized the European Centre for Medium-Range Weather Forecasts (ECMWF) Reanalysis v5 (ERA5) dataset. A total of 66 daily mean atmospheric variables, aggregated from hourly data were used. These comprised six single-level variables and five variables sampled at 12 pressure levels, extending from near the surface up to the lower stratosphere. These include essential prognostic fields such as zonal and meridional wind components, temperature, humidity, and geopotential, which
are crucial to define atmospheric dynamics. In addition, orography was included as a static topographic variable, resulting in a total of 67 input channels. The dataset was processed into tensors of shape 72 × 144 × 67, corresponding to a 2.5° spatial resolution. For model training and evaluation, the dataset was split into a training
 (1979–2015), validation (2016–2017), and test (2018) periods.

\begin{table}[h!]
\centering
\caption{Overview of meteorological variables used from the ERA5 dataset, including pressure-level and surface-level fields.}
\label{tab:variables}
\begin{tabular}{llll}
\toprule
\textbf{Variable name} & \textbf{Short name} & \textbf{Vertical levels (hPa)} & \textbf{Units} \\
\midrule
\multicolumn{4}{l}{\textit{Pressure-level variables}} \\
Zonal wind           & U     & 1000, 925, 850, 800,         & m/s     \\
                     &       & 700, 600, 500, 400,          &         \\
                     &       & 300, 200, 100, 50            &         \\
Meridional wind      & V     & same as above                & m/s     \\
Temperature          & T     & same as above                & K       \\
Specific humidity    & Q     & same as above                & kg/kg   \\
Geopotential         & Z     & same as above                & m\textsuperscript{2}/s\textsuperscript{2} \\
\midrule
\multicolumn{4}{l}{\textit{Single-level variables}} \\
2m temperature               & T2m   & -     & K          \\
Mean sea level pressure      & MSL   & -     & Pa         \\
Surface air pressure         & SP    & -     & Pa         \\
Total column water vapor     & TCWV  & -     & kg/m\textsuperscript{2} \\
Skin temperature             & SKT   & -     & K          \\
TOA incident solar radiation & TISR  & -     & J/m\textsuperscript{2} \\
\bottomrule
\end{tabular}
\end{table}

\subsection{Overall architecture of KAI-$\alpha$}
\begin{figure}[h!]
	\centering
	\includegraphics[width=0.9\textwidth]{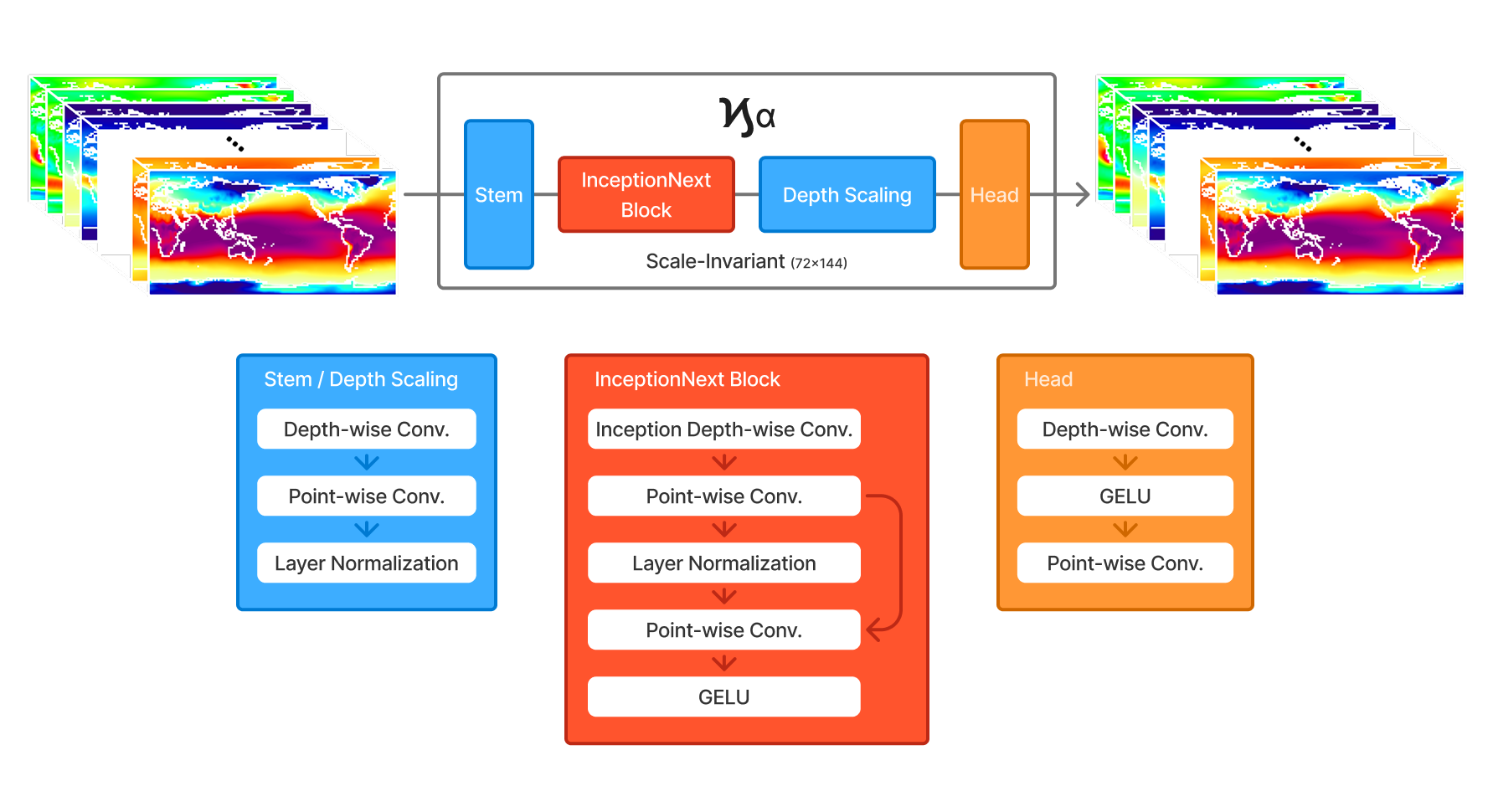}
	\caption{Overall architecture of the proposed KAI-$\alpha$ model for global weather forecasting. The model takes as input a set of 67 atmospheric variables on a 72×144 grid and processes them through a streamlined convolutional pipeline. The design includes three main components: Stem, InceptionNeXt Blocks with Depth Scaling, and the Head. Each module incorporates lightweight yet expressive operations such as depthwise/pointwise convolutions and LayerNorm. The InceptionNeXt block introduces geophysically-aware depthwise convolution, followed by GELU activation for non-linearity.}
	\label{fig:kai_model}
\end{figure}

Our neural network architecture is a deep convolutional model inspired by InceptionNeXt, and its overall structure is described in Figure \ref{fig:kai_model}. Initially, the data are embedded using a stem block that employs a depthwise separable convolution combined with layer normalization, setting the stage for multi-scale feature extraction. This is followed by a series of depth scaling layers and four hierarchical stages, each comprising a sequence of residual blocks. Each block integrates an Inception-style depthwise convolution module with specialized Geocyclic padding to handle the geospatial periodicity, as well as pointwise convolutions for efficient channel mixing. Unlike typical encoder-decoder or U-shaped architectures, our model maintains a scale-invariant structure without upsampling paths, focusing instead on efficient spatial and channel-wise representation learning. The InceptionNeXt blocks are repeated in a {3, 3, 15, 3} configuration with corresponding channel dimensions of {48, 96, 192, 288}. For the final head, a depthwise separable convolution is again employed to extract refined features and adjust the output to match the desired number of predicted channels.

\subsubsection{InceptionNeXt}
InceptionNeXt is a convolutional neural network that refines the conventional large‐kernel depthwise convolution by decomposing it into four parallel branches along the channel dimension: a small square 3×3 convolution branch, two orthogonal band convolution branches (one with 1×11 kernel and the other with 11×1 kernel) designed to capture extended spatial dependencies efficiently, and an identity mapping branch that leaves a subset of channels unchanged. These branches are subsequently concatenated and normalized, then processed within a residual framework to form a streamlined token mixer. This design is integrated within a multi-stage architecture characterized by progressive downsampling and increasing channel dimensions, which together yield a superior accuracy–speed trade-off, as evidenced in both image classification and semantic segmentation tasks. Moreover, the use of band convolutions is particularly advantageous for datasets like ERA5, as their large receptive fields enable the efficient capture of extensive atmospheric phenomena, thereby enhancing the network's ability to model complex, large-scale patterns \cite{yu2024inceptionnext}. 

\subsection{Modernizing CNN for global weather forecast}
In the following sections, we outline a systematic roadmap for advancing from a baseline model to an enhanced convolutional network with competitive performance. In this study, the model proposed by \cite{weyn2020improving}, which features a CNN architecture with a U-Net-like structure, serves as the baseline. Building upon this foundation, we introduce a stepwise modification classified as micro, macro, and meta designs.

\subsubsection{Macro Design}
The macro design of our architecture refers to high-level structural decisions that govern the overall information flow and spatial encoding across the network. These choices define how spatial features are extracted, aggregated, and processed throughout the model. In our design, we introduce several key modifications aimed at improving performance while maintaining computational efficiency.

We first incorporate a stem layer designed to enhance feature extraction at the input stage. Specifically, we apply a depthwise separable convolution followed by layer normalization to efficiently extract low-level spatial features. 

To enable deeper hierarchical feature extraction, we adopt a staged architecture with four progressively deepening stages. Each stage increases both the network depth and feature dimensionality to capture multi-scale atmospheric patterns. The architecture features a repetition pattern of [3, 3, 15, 3] with feature dimensions [48, 96, 192, 288] as proposed by \cite{yu2024mambaout}. The first two stages replicate the main block three times, followed by channel expansion through downsampling. The third stage, which is the deepest, applies 15 block repetitions at 192 channels before transitioning to 288 channels for the final stage. \cite{yu2024mambaout}. 

Further improvements are achieved by replacing the main blocks with inverted residual blocks incorporating pointwise convolution. In this structure, the input is first expanded to a higher-dimensional space using a pointwise convolution, followed by a non-linear transformation, and then projected back to the original dimension through another pointwise convolution. A residual connection is maintained between the shallow input and output to preserve low-level features. This design improves efficiency by limiting computationally expensive operations to the expanded feature space, achieving a better balance between capacity and cost \cite{sandler2018mobilenetv2}.

\subsubsection{Micro Design}
The micro design focuses on localized architectural components that govern the behavior of individual layers, such as activation functions and normalization techniques. These components play a critical role in the expressiveness and stability of the network. In our model, we replace the modified Leaky ReLU used in the original DLWP framework with the Gaussian Error Linear Unit (GELU), a smoother and more adaptive activation function \cite{hendrycks2016gaussian} \cite{weyn2020improving}.

\subsubsection{Meta Design}

The meta design of our model is grounded in the architectural decisions that align with the characteristics of Earth’s spherical geometry and atmosphere system. Two key components define this design: geocyclic padding and a scale-invariant structure.
Geocyclic padding addresses distortions that arise when processing latitude-longitude gridded data by ensuring spatial continuity across longitudinal boundaries and poles \cite{cheon2024karina}. Specifically, circular padding is applied along the longitudinal axis, and pole-side reordering is performed to prevent artificial discontinuities at the dateline and high-latitude boundaries. This approach is essential for global atmospheric reanalysis datasets such as ERA5, which are provided on a cylindrical map projection where physical continuity is inherent but broken by standard padding schemes.

Introducing a scale-invariant model structure is another substantial change. Rather than following traditional U-Net–style downsampling and upsampling, the original resolution is maintained throughout the entire network to avoid information loss from aggressive pooling operations. While this approach considerably increases calculational and memory demands, the low spatial resolution setup at the  2.5° grids, combined with the extended receptive fields provided by InceptionNeXt, enables effective capture of large-scale phenomena and teleconnection patterns in global atmospheric fields with affordable computational cost. We consider this is a parsimonious design that functions as a pseudo-global attention algorithm.

\subsection{Training Details}
To ensure that the model properly considers the characteristics of the ERA5’s cylindrical map projection, we adopted a latitude-weighted root mean square error (weighted RMSE) as the loss function. Unlike standard RMSE, this approach assigns greater importance to errors at lower latitudes, where grid size is larger due to the spherical nature of the Earth. The weighting factor is derived from the cosine of latitude, ensuring that each latitude band contributes proportionally to its area. The computation of the weighted RMSE follows the formulation:

\begin{equation}
A(\varphi_i) = N_{\text{lat}} \frac{\cos \varphi_i}{\sum\limits_{l=1}^{N_{\text{lat}}} \cos \varphi_l}
\end{equation}

where $N_{\text{lat}}$ represents the number of grid points in the latitudinal direction.

The RMSE$_l$ between ground truth (i.e., ERA5) $y_{t,i,j}$ and forecast $\hat{y}_{t,l,i,j}$ for lead time $l$ is defined as follows:

\begin{equation}
\text{RMSE}_l = \sqrt{
\frac{1}{N_{\text{time}} N_{\text{lat}} N_{\text{lon}}}
\sum_{t=1}^{N_{\text{time}}}
\sum_{i=1}^{N_{\text{lat}}}
\sum_{j=1}^{N_{\text{lon}}}
A(\varphi_i)(\hat{y}_{t,l,i,j} - y_{t,i,j})^2}
\end{equation}

For training, no additional fine-tuning was performed. The model was trained for about 12 hours using a single NVIDIA L40s GPU. The training process leveraged mixed precision to optimize memory efficiency while maintaining numerical stability. 

Table \ref{tab:training_time_gpus} compares the training configurations of several state-of-the-art AI-based weather forecasting models. It highlights the number of GPUs or TPUs used, the specific hardware version, and the total training time required for each model. Notably, large-scale models like Pangu Weather and GraphCast required massive computational resources—up to 192 GPUs or multi-week training durations. In contrast, KAI-$\alpha$ demonstrates high training efficiency by achieving comparable performance with only 1 Nvidia L40s GPU in 12 hours, underscoring its significantly lightweight design. 

Table \ref{tab:training} outlines the hyperparameter configuration used to train the KAI-$\alpha$ model on the daily ERA5 dataset. The model is trained using a time step (dt) of 1 day. The learning rate is set to 0.001, with a CosineAnnealingLR scheduler applied to gradually reduce the learning rate over training. The model uses Z-score normalization and takes in and predicts 67-channel inputs and outputs, corresponding to multiple atmospheric variables. AdamW is used as the optimizer for stable convergence, and the model is trained for a maximum of 150 epochs. These settings were chosen to ensure balanced learning dynamics and convergence stability over long-range forecasts.

\begin{table}[h!]
\centering
\caption{Comparison of training configurations and computational cost for recent AI-based weather forecasting models.}
\label{tab:training_time_gpus}
\begin{tabular}{@{}lccc@{}}
\toprule
\textbf{Model} & \textbf{Hardware Configuration} & \textbf{Training Time} & \textbf{V100-Equivalent GPU-Days} \\
\midrule
AIFS \cite{lang2024aifs}               & 64 Nvidia A100 GPUs      & 1 week     & $\sim$1,344.0 \\
ArchesWeather-M \cite{couairon2024archesweather} & 2 Nvidia A100 GPUs & 2.5 days   & $\sim$15.0 \\
FuXi \cite{chen2023fuxi}               & 8 Nvidia A100 GPUs       & 30 hrs     & $\sim$10.0 \\
FengWu \cite{chen2023fengwu}           & 32 Nvidia A100 GPUs      & 17 days    & $\sim$544.0 \\
FourCastNet \cite{pathak2022fourcastnet} & 64 Nvidia A100 GPUs    & 16 hrs     & $\sim$42.7 \\
GraphCast \cite{lam2022graphcast}      & 32 Google TPU v4 devices & 4 weeks    & $\sim$896.0* \\
Pangu-Weather \cite{bi2023accurate}    & 192 Nvidia V100 GPUs     & 64 days    & 12,288.0 \\
Stormer \cite{nguyen2024scaling}       & 128 Nvidia A100 GPUs     & 24 hrs     & $\sim$426.7 \\
\midrule
\textbf{KAI-$\boldsymbol{\alpha}$}     & \textbf{1 Nvidia L40s GPU} & \textbf{12 hrs} & \textbf{$\sim$2.4} \\
\bottomrule
\end{tabular}
\end{table}

\begin{table}[h!]
\centering
\caption{Training hyperparameters used for KAI-$\alpha$ model on the ERA5 dataset.}
\label{tab:training}
\begin{tabular}{ll}
\toprule
\textbf{Hyperparameter} & \textbf{Value} \\
\midrule
Loss & L2 \\
Learning Rate (LR) & 0.001 \\
Scheduler & CosineAnnealingLR \\
dt & 1 day \\
Number of In-Channels & 67 \\
Number of Out-Channels & 67 \\
Normalization & Z-score \\
Optimizer Type & AdamW \\
Max Epochs & 150 \\
\bottomrule
\end{tabular}
\end{table}

\section{Result}
\subsection{Performance Gains from CNN Modernization in Global Forecast Tasks}
\subsubsection{Macro Design}
\begin{figure}[h!]
	\centering
	\includegraphics[width=\textwidth]{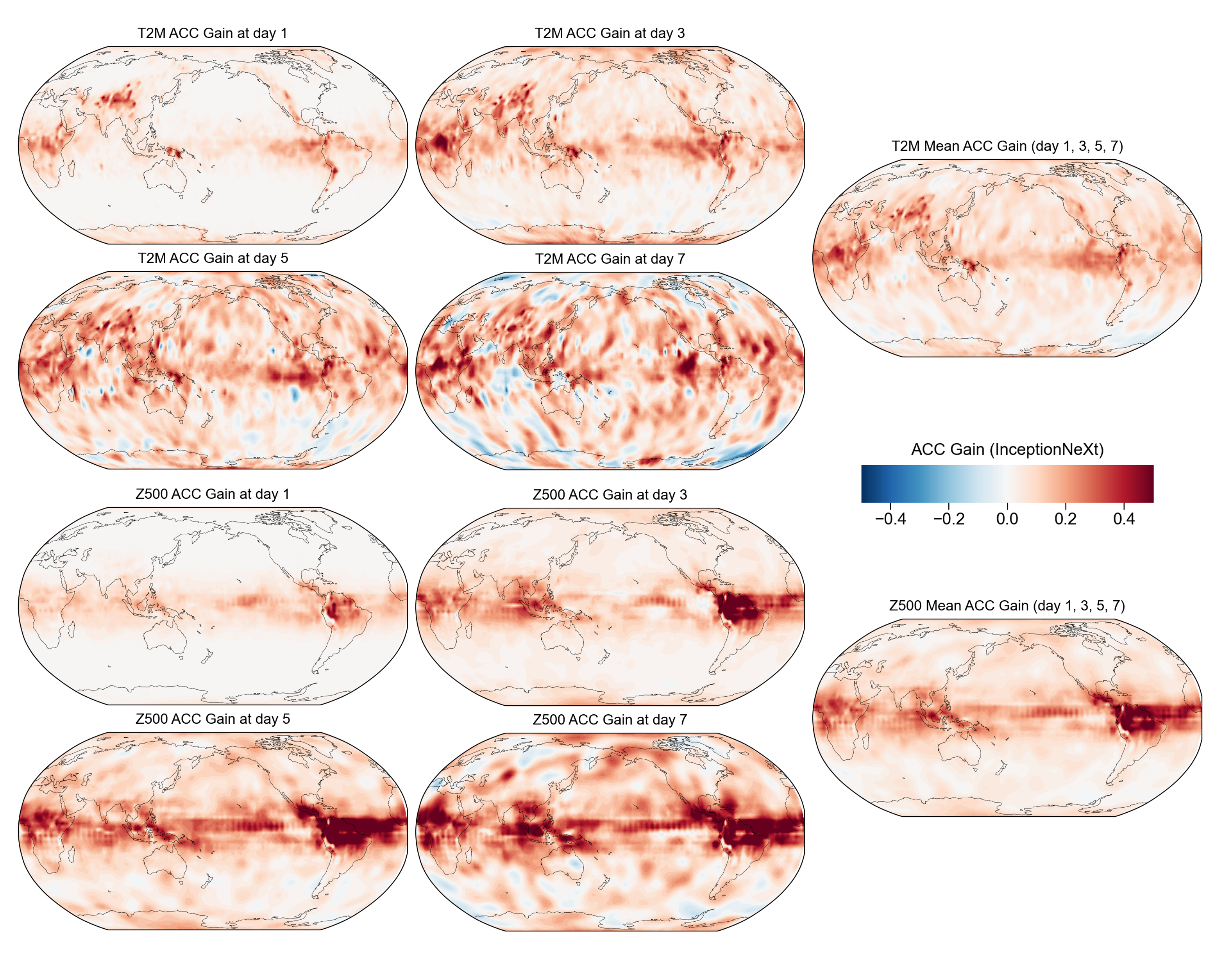}
	\caption{Spatial distribution of ACC gain from InceptionNeXt over baseline model for T2M and Z500 forecasts at lead times of 1, 3, 5, and 7 days. Each panel shows the anomaly correlation coefficient (ACC) improvement across the globe, with warm colors indicating performance gains and cool colors indicating reductions. The rightmost panels present the multi-day mean ACC gain (days 1, 3, 5, 7) for T2M (top) and Z500 (bottom), highlighting consistent skill improvements, especially in equatorial and midlatitude regions.}
	\label{fig:spatial}
\end{figure}

Macro design improves the model’s capacity for spatial encoding without significantly increasing computational complexity, resulting in an ACC improvement from 0.425 to 0.500 on the 500 hPa geopotential height (Z500) at a 7-day forecast lead. This design remains lightweight with a total cost of 2.26 GFLOPs, only marginally higher than the 2.41 GFLOPs of the Weyn et al. (2020) baseline. The InceptionNeXt structure enhances the network’s ability to recognize complex global features, leading to an ACC increase from 0.500 to 0.552, with an associated increase in computational cost to 25.16 GFLOPs. As described in Figure \ref{fig:spatial}, the ACC gains for both 2-meter temperature (T2M) and 500 hPa geopotential height (Z500) are particularly concentrated around the equatorial region.
Furthermore, incorporating pointwise convolution with inverted residual blocks improves efficiency by limiting computationally expensive operations to the expanded feature space, achieving a better balance between capacity and cost. As a result, the model achieves an ACC improvement from 0.552 to 0.576, with a modest increase in GFLOPs from 25.16 to 29.59.

\subsubsection{Micro Design}
When applied to the macro design configuration, the use of GELU leads to an ACC improvement from 0.644 to 0.664 for Z500 prediction at a 7-day lead, without adding any additional GFLOPs (remaining at 10.80 GFLOPs). Furthermore, when GELU is applied on top of the full meta design—which includes geocyclic padding and scale-invariant layers—the ACC further improves from 0.692 to 0.721. This indicates that GELU consistently enhances performance, not only at the micro level but also when combined with broader architectural innovations, which we detail in the following section.

\subsubsection{Meta Design}
Incorporating Geocyclic padding alone results in a modest performance gain, increasing the ACC score from 0.644 to 0.646 for Z500 prediction at day-7 lead, without adding any additional computational cost (remaining at 10.80 GFLOPs). The scale-invariant design enables more stable and spatially consistent feature representations across the globe. Implementing this structure leads to a significant ACC improvement from 0.646 to 0.692, but increases the total computational cost to 156.72 GFLOPs, due to the higher spatial fidelity preserved across all layers. Together, these meta design components ensure that the model architecture is well-aligned with the physical properties of global climate data, allowing for more accurate and stable long-range forecasting. Although the final KAI-$\alpha$ model exhibits higher FLOPs due to its enhanced meta design, it remains highly efficient in practice, requiring only 12 hours of training on a single NVIDIA L40s GPU. 

\begin{figure}[h!]
	\centering
	\includegraphics[width=\textwidth]{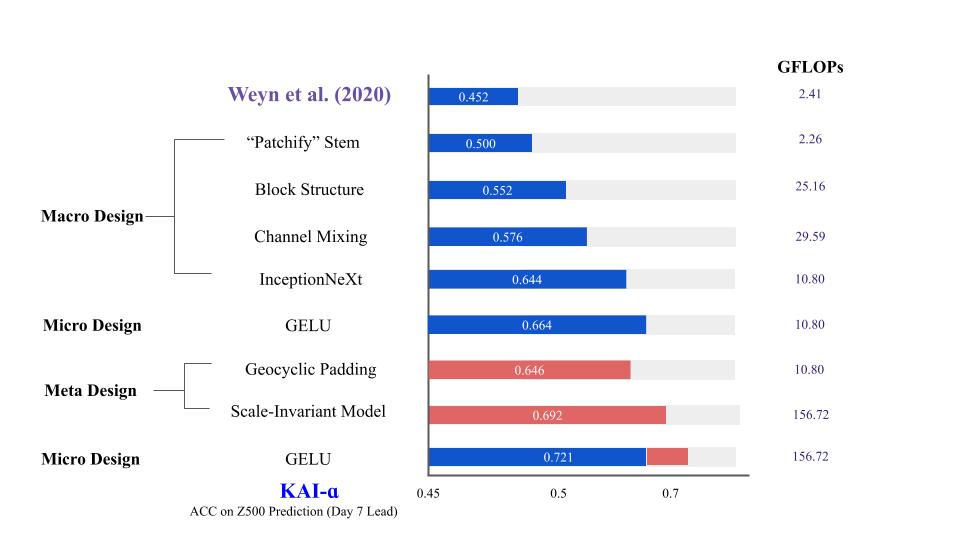}
	\caption{Ablation study of macro, micro, and meta design components in KAI-$\alpha$, evaluated by Z500 forecast skill at day-7 lead time. The bar chart illustrates the contribution of each architectural design choice to the anomaly correlation coefficient (ACC) on Z500 prediction at day 7. Design stages are grouped into macro design (stem, block structure, channel mixing, InceptionNeXt), micro design (activation function), and meta design (geocyclic padding, scale-invariant structure). The pink bars highlight the performance gain from meta design components. The final KAI-$\alpha$ model combines all components, with the rightmost pink segment indicating the additional improvement achieved beyond the macro design baseline.}
	\label{fig:fig1}
\end{figure}

The entire architectural design process, encompassing macro, micro, and meta components, along with their respective performance contributions, is summarized in Figure~\ref{fig:fig1}.

\subsection{Comparison with SOTA models}
\begin{figure}[h!]
	\centering
	\includegraphics[width= 0.8\textwidth]{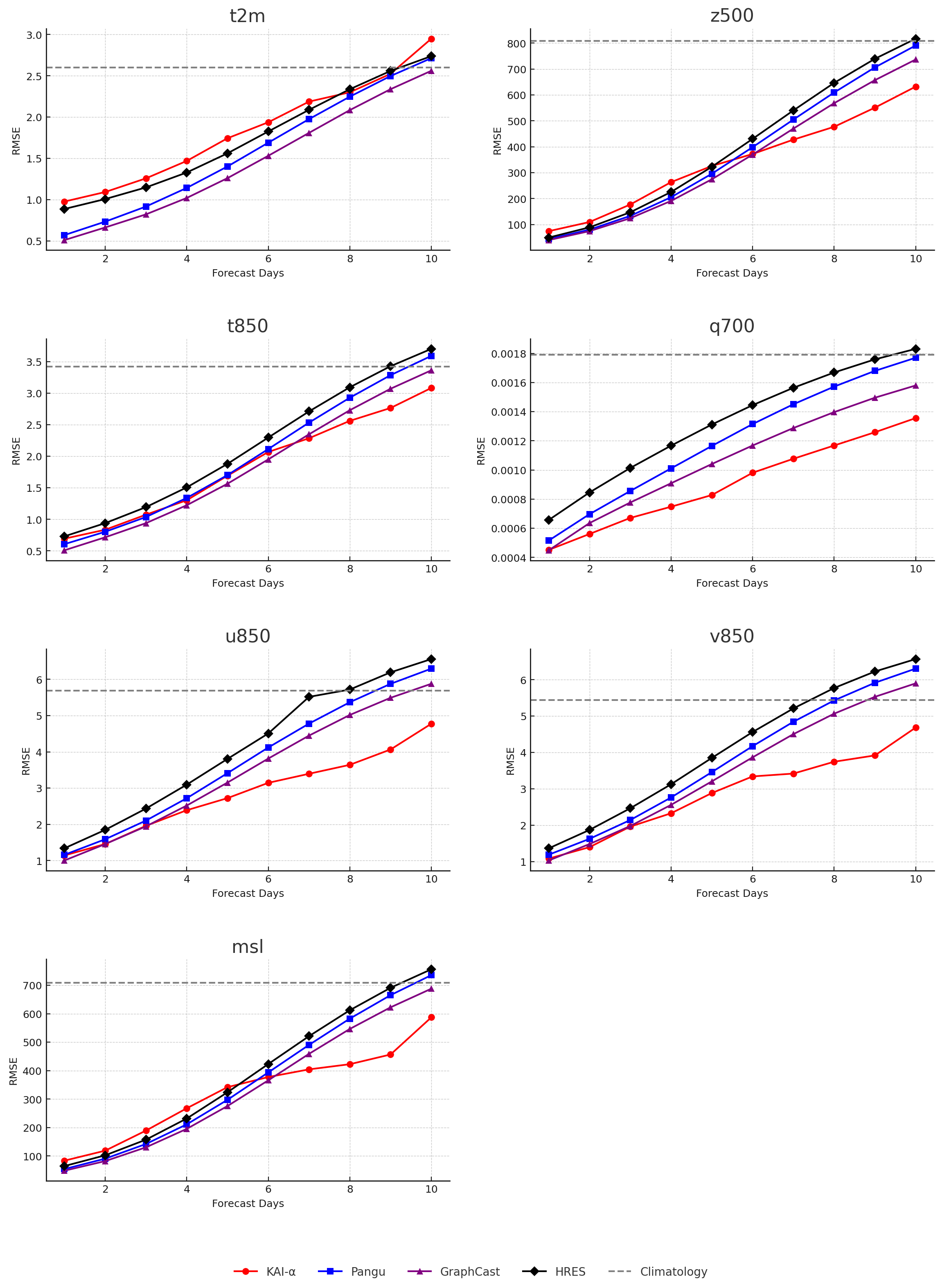}
	\caption{Root Mean Square Error (RMSE) over 10 forecast days for key atmospheric variables, comparing KAI-$\alpha$, Pangu, GraphCast, and HRES. Each panel shows the RMSE progression over time for one of the following variables: T2M, Z500, T850, Q700, U850, V850, and MSLP. The proposed KAI-$\alpha$ model (red) consistently achieves lower RMSE than other models, particularly at longer lead times.}
	\label{fig:kai_sota}
\end{figure}

We evaluate the forecast performance of KAI-$\alpha$ in comparison with two state-of-the-art (SOTA) deep learning-based models, Pangu-Weather and GraphCast, both available through the WeatherBench-2 benchmark \cite{rasp2024weatherbench}. All models are assessed using the 2018 ERA5 reanalysis dataset. Since KAI-$\alpha$ is trained at a 2.5° resolution, bilinear interpolation is applied to align its outputs with the 1.5° target resolution used for evaluation. Additionally, we include the IFS HRES operational forecast system as a reference. For consistency, all model outputs are evaluated at a common resolution of 1.5°. Figure \ref{fig:kai_sota} shows the Root Mean Square Error (RMSE) over 10 forecast days for key atmospheric variables: T2M, Z500, T850, Q700, U850, V850, and MSLP. Across nearly all variables and lead times, KAI-$\alpha$ exhibits lower RMSE than the other models. The performance advantage is particularly evident at longer lead times (day 7–10), where KAI-$\alpha$ shows reduced error growth compared to Pangu-Weather, GraphCast, and HRES. The gray dashed lines represent the RMSE of climatology, and KAI-$\alpha$ remains below this baseline across all variables throughout the forecast horizon. These results confirm that KAI-$\alpha$ achieves robust and competitive forecasting skill across a range of dynamical and thermodynamic fields. 

Unlike the RMSE, which is evaluated at 1.5° after interpolation, we compute the Anomaly Correlation Coefficient (ACC) score based on the original 2.5° resolution dataset to preserve consistency with the training resolution of KAI-$\alpha$. The ACC metric is calculated using the following equation:

\begin{equation}
\text{ACC}_l = \frac{1}{N_{\text{time}}} \sum_{t=1}^{N_{\text{time}}}
\frac{
\sum_{i=1}^{N_{\text{lat}}} \sum_{j=1}^{N_{\text{lon}}} A(\varphi_i)(\hat{y}_{t,l,i,j} - \overline{\hat{y}_{t,l}})(y_{t,i,j} - \overline{y_t})
}{
\sqrt{
\sum_{i=1}^{N_{\text{lat}}} \sum_{j=1}^{N_{\text{lon}}} A(\varphi_i)(\hat{y}_{t,l,i,j} - \overline{\hat{y}_{t,l}})^2
\sum_{i=1}^{N_{\text{lat}}} \sum_{j=1}^{N_{\text{lon}}} A(\varphi_i)(y_{t,i,j} - \overline{y_t})^2
}
}
\end{equation}

\begin{figure}[h!]
	\centering
	\includegraphics[width= 0.8\textwidth]{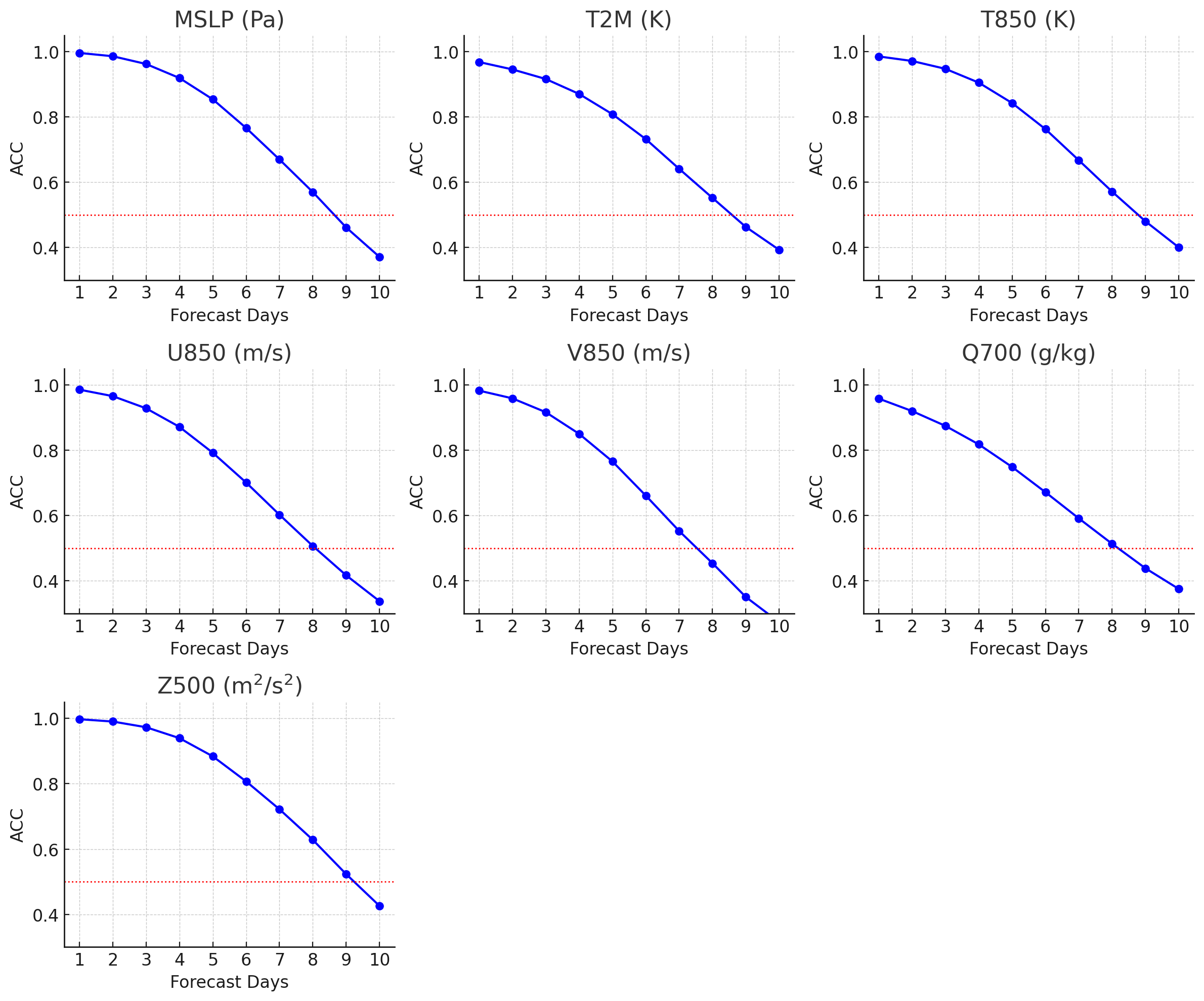}
	\caption{Anomaly Correlation Coefficient (ACC) over 10 forecast days for key atmospheric variables.Forecast skill is shown for MSLP, T2M, T850, U850, V850, Q700, and Z500, with the red dashed line indicating the ACC threshold of 0.5}
	\label{fig:kai_acc}
\end{figure}

Figure \ref{fig:kai_acc} illustrates the forecast skill of the KAI-$\alpha$ model across 10 forecast days, evaluated using the Anomaly Correlation Coefficient (ACC) for seven key atmospheric variables. The plots show a gradual decline in ACC with increasing lead time, which is expected in numerical forecasting. Notably, all variables maintain ACC values above the skill threshold of 0.5 (indicated by the red dashed line) up to approximately day 7–9, depending on the variable.

\subsection{Ablation Studies}
We analyze the effect of three different channel shuffling methods including using pointwise convolution, MLP, and ConvMLP \cite{trockman2022patches}\cite{tolstikhin2021mlp}\cite{yu2024inceptionnext}. Pointwise convolution is an operation that mixes information across channels while preserving spatial resolution, by using 1x1 convolution. MLP typically applies fully connected layers to each spatial location independently, often requiring spatial flattening, which can lead to a loss of spatial structure. In contrast, ConvMLP replaces the fully connected layers with 1×1 convolutions, enabling spatially-aware channel mixing while maintaining the spatial dimensions of the input. 

\begin{figure}[h!]
	\centering
	\includegraphics[width= 0.9\textwidth]{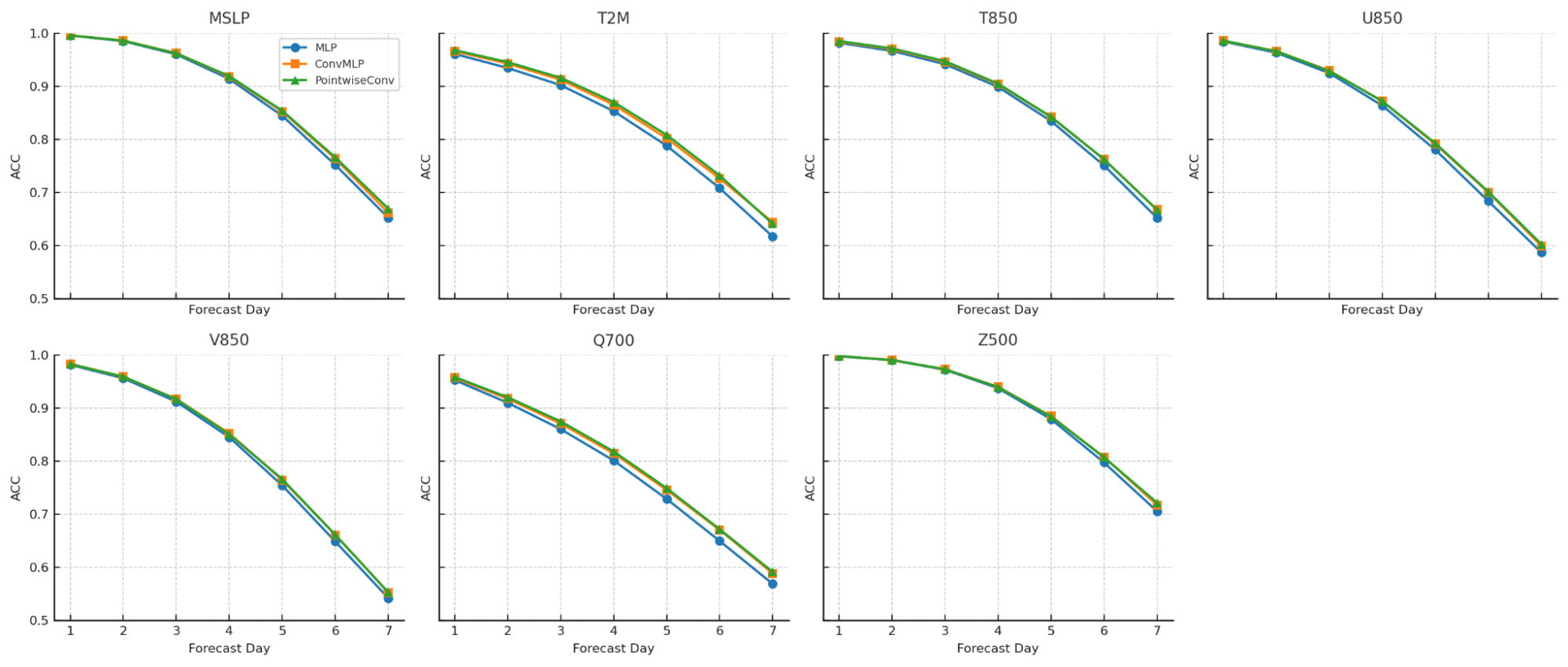}
	\caption{Ablation study comparing MLP, Conv1D, and ConvMLP for channel mixing across forecast days. Each panel shows the forecast accuracy (pattern correlation) over 7 days for key atmospheric variables: MSLP, T2M, T850, U850, V850, Q700, and Z500.}
	\label{fig:kai_ablation}
\end{figure}

To evaluate the effectiveness of different channel-mixing strategies in the  KAI-$\alpha$ model, we compared MLP, ConvMLP, and PointwiseConv modules across seven key meteorological variables (e.g., MSLP, T2M, U850) over a 7-day forecast. As shown in Figure \ref{fig:kai_ablation}, all three methods exhibit similar trends in ACC degradation as the forecast day increases, with PointwiseConv and ConvMLP consistently outperforming the standard MLP, especially beyond Day 4. Between ConvMLP which is used in InceptionNeXt, and PointwiseConv, the performance difference was marginal across all variables and forecast ranges. However, ConvMLP introduces additional convolutional layers, leading to an increased number of parameters and computational overhead. With the negligible performance gap but clear computational advantage, we adopted PointwiseConv as the final channel-mixing method for the  KAI-$\alpha$ model. This decision allows the model to maintain high forecast accuracy while remaining lightweight and efficient, which is especially critical for deployment in climate and weather forecasting systems

\FloatBarrier 

\subsection{Case Study (European Heatwave)}
\begin{figure}[h!]
	\centering
	\includegraphics[width=0.8\textwidth]{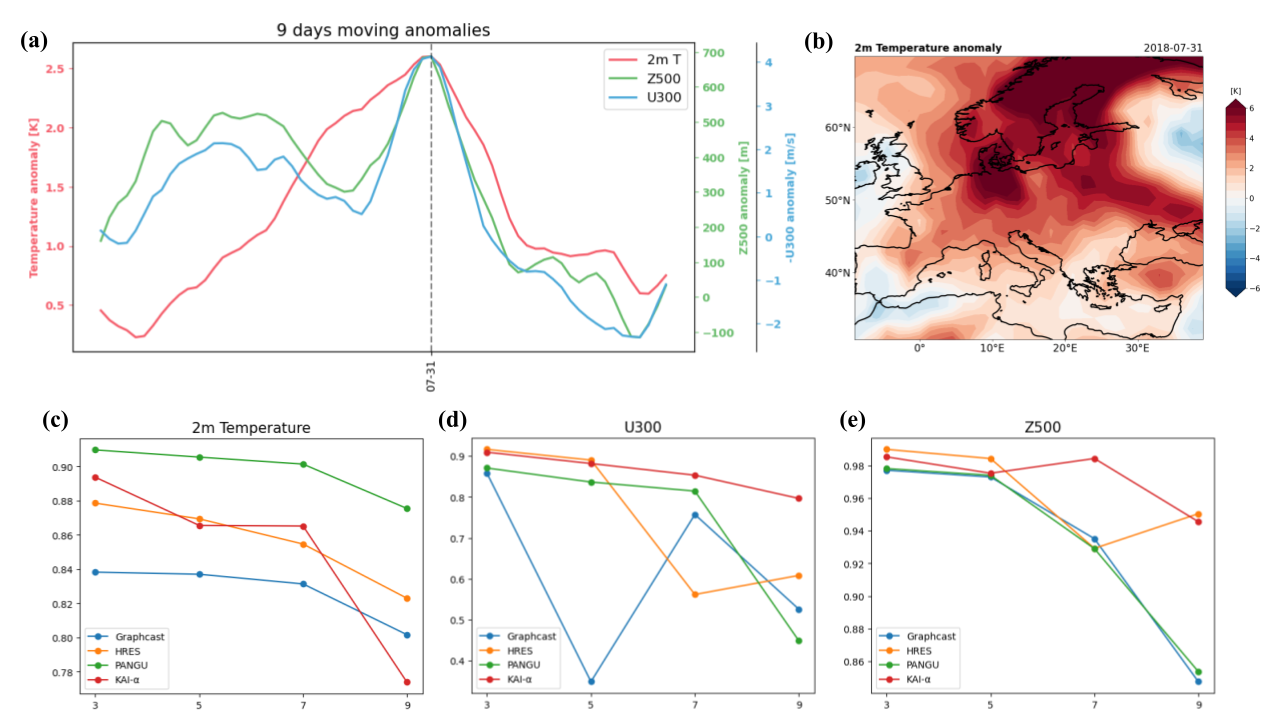}
    ....................
	\caption{(a) Temporal evolution of 9-day moving anomalies for T2M, Z500, and U300 during the 2018 European heatwave. (b) Spatial distribution of surface temperature anomalies at the peak of the event. (c–e) Pattern correlation performance of GraphCast, HRES, PANGU, and KAI models across forecast lead times for T2M, U300, and Z500, respectively.}
	\label{fig:kai_heatwave}
\end{figure}

In the summer of 2018, northwestern Europe experienced an unusually severe heatwave and drought \cite{rousi2023extremely} \cite{knutzen2025impacts} \cite{salomon20222018} \cite{bastos2020direct} \cite{peters2020historical}. This heatwave was attributed to a persistent high-pressure blocking system and a weakened jet stream, both of which contributed to prolonged heat accumulation near the surface. Figure~\ref{fig:kai_heatwave} provides a comprehensive view of this extreme event and the associated forecasting model performances. Panel (a) illustrates the 9-day moving anomalies of 2-meter temperature (T2M), 500 hPa geopotential height (Z500), and upper-tropospheric zonal wind (U300). The concurrent peaks in all three variables during late July confirm the intensification of the blocking system and the suppressed jet stream that were central to the heatwave's development. Panel (b) depicts the spatial anomaly of T2M during the peak of the event, showing extreme positive temperature anomalies across northwestern and central Europe. This spatial structure is consistent with previous studies linking the heatwave to mid-tropospheric ridging and disrupted jet stream flow \cite{rousi2023extremely}. Panels (c–e) compare the pattern correlation skill of four models—GraphCast, HRES, PANGU, and KAI-$\alpha$—for T2M, U300, and Z500 at forecast lead times of 3, 5, 7, and 9 days. Notably, KAI-$\alpha$ demonstrates consistently high pattern correlation, particularly for Z500 and U300, indicating its strong capacity to capture the large-scale dynamical drivers of the compound extreme event, including the evolution of the blocking structure and weakened jet stream.

\subsection{Case Study (East Asian Monsoon)}
\begin{figure}[h!]
	\centering
	\includegraphics[width=0.8\textwidth]{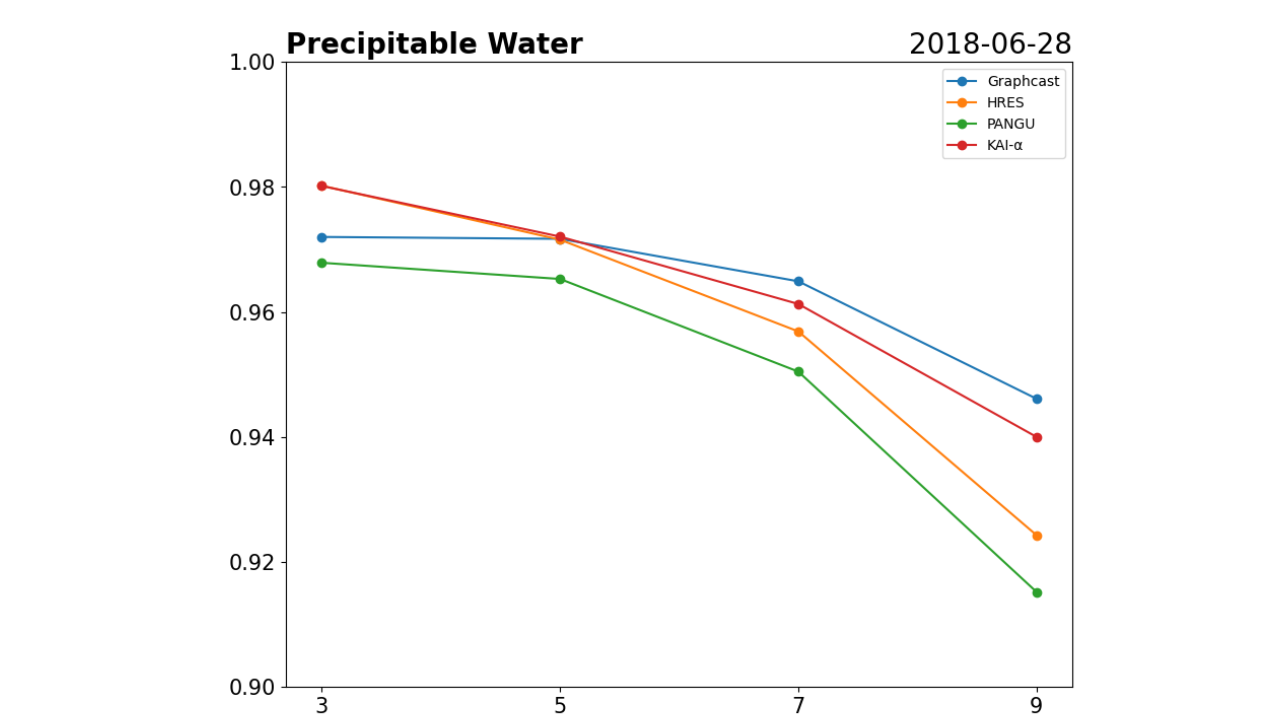}
	\caption{(a) Pattern correlation of precipitable water over the East Asian region (120°E–180°, 0–60°N) on 28 June 2018. Correlations are computed between ERA5 and forecasts from Graphcast (blue), HRES (orange), PANGU (green), and KAI-$\alpha$ (red) at lead times of 3, 5, 7, and 9 days.}
	\label{fig:kai_precipitable_corr}
\end{figure}

On 28 June 2018, the East Asian monsoon front was positioned near the East Asian continent, and the tropical cyclone Prapiroon was approaching from the equatorial region. Under these conditions, we evaluate the performance of the KAI-$\alpha$ model in simulating precipitable water, in comparison with other existing models; Graphcast, PANGU, and HRES. Figure~\ref{fig:kai_precipitable_corr}  prsents the pattern correlation of precipitable water forecasts from four models for 28 June 2018, across lead times of 3, 5, 7, and 9 days. At a 3-day lead time, HRES and KAI-$\alpha$ showed the highest correlation (~0.98). As lead time increased, overall performance declined. Among the four models, Graphcast maintained the highest correlation beyond day 7, followed by KAI-$\alpha$.

\begin{figure}[htbp]
	\centering
	\includegraphics[width=0.8\textwidth]{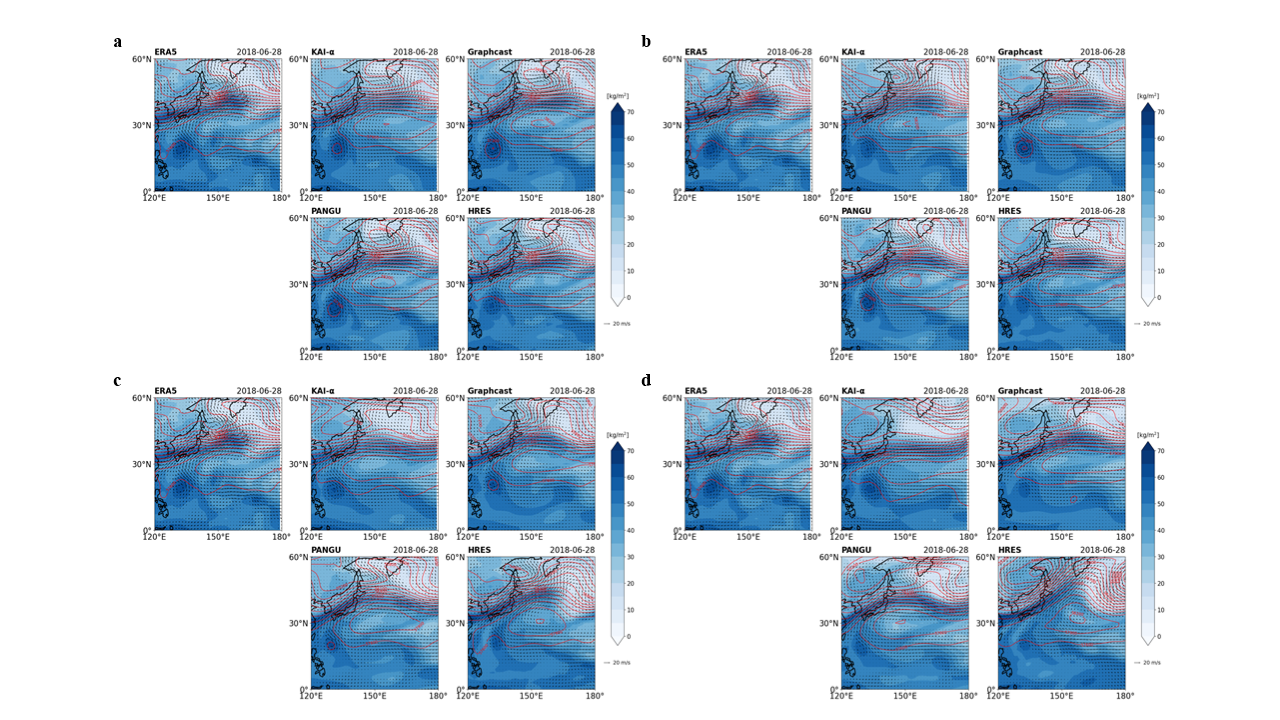}
	\caption{Spatial distribution of precipitable water (shading; $\mathrm{kg}\,\mathrm{m}^{-2}$), 500 hPa geopotential height (contours; 300 m interval), and 850 hPa wind vectors over the East Asian region (120°E–180°, 0–60°N) on 28 June 2018. Panels (a)–(d) correspond to forecasts with lead times of 3, 5, 7, and 9 days, respectively. Results from ERA5, KAI-$\alpha$, Graphcast, PANGU, and HRES are presented for each lead time.}
	\label{fig:kai_precip}
\end{figure}

To further assess the model ability to simulate East Asian summer weather systems, PW, 500 hPa geopotential height, and 850 hPa wind fields were compared in lead times (Figure \ref{fig:kai_precip}). At a 3-days lead time (Figure 9a), both KAI-$\alpha$ and HRES showed high pattern similarity not only in precipitable water but also in the 500 hPa geopotential height field, particularly north of the stationary front. In terms of the tropical cyclone, HRES provided the most accurate representation of the cyclone's pressure structure, whereas KAI-$\alpha$, PANGU, and Graphcast tended to overestimate its intensity. At a 5-days lead time (Figure 9b), KAI-$\alpha$ and HRES continued to capture the position of the stationary front and the associated circulation patterns to the north and south of it, including the tropical cyclone. By the 7-days lead time (Figure 9c), both KAI-$\alpha$ and Graphcast successfully reproduced the geopotential height field north of the front, with KAI-$\alpha$ more accurately capturing the location and intensity of the tropical cyclone. At the longest lead time of 9 days (Figure 9d), Graphcast better simulated the geopotential field over northern East Asia and the high-pressure system south of the front, while KAI-$\alpha$ produced a more realistic representation of the tropical cyclone, which was displaced too far north in Graphcast compared to ERA5. Overall, KAI-$\alpha$ demonstrated strong performance across all lead times, particularly in accurately predicting the location and intensity of the tropical cyclone in long-range forecasts.

\section{Conclusion}
This paper introduces KAI-$\alpha$, a lightweight convolutional architecture designed for global weather forecasting. The model builds on Weyn et al.’s CNN-based approach and incorporates a systematic design roadmap. Key architectural enhancements include scale-invariant modeling and InceptionNeXt-based blocks, along with design choices that account for the spatial structure of the Earth—each contributing to improved performance and computational efficiency. Trained on the ERA5 daily dataset with only about 7 million parameters, KAI-$\alpha$ delivers competitive accuracy in medium-range forecasts, particularly beyond 7 days. Our case studies, including the 2018 European heatwave and the East Asian monsoon precipitable water events, demonstrate KAI-$\alpha$'s robust capacity to accurately capture large-scale dynamical drivers and tropical cyclone evolution, maintaining high pattern correlation even at longer lead times. Our findings demonstrate that well-designed convolutional models can serve as strong alternatives to large Transformer-based systems. This highlights their potential as an efficient and practical solution for global weather prediction.

\bibliographystyle{unsrtnat}
\bibliography{references}
\clearpage 

\section{Supplementary Materials}
We visualize forecasts produced by our model at lead times ranging from 1 day to 14 days for seven key atmospheric variables, including temperature, humidity, wind, and pressure-related fields. All forecasts are initialized at 00 UTC on January 1st, 2018, a representative wintertime case. Each panel corresponds to a specific lead time, and each row represents a different variable. The columns show (1) the initial condition at the start time, (2) the ground truth at the target lead time, (3) the model-generated forecast, and (4) the forecast error (bias), computed as the difference between the forecast and the ground truth. 

\begin{figure}[htbp]
	\centering
	\includegraphics[height=0.85\textheight]{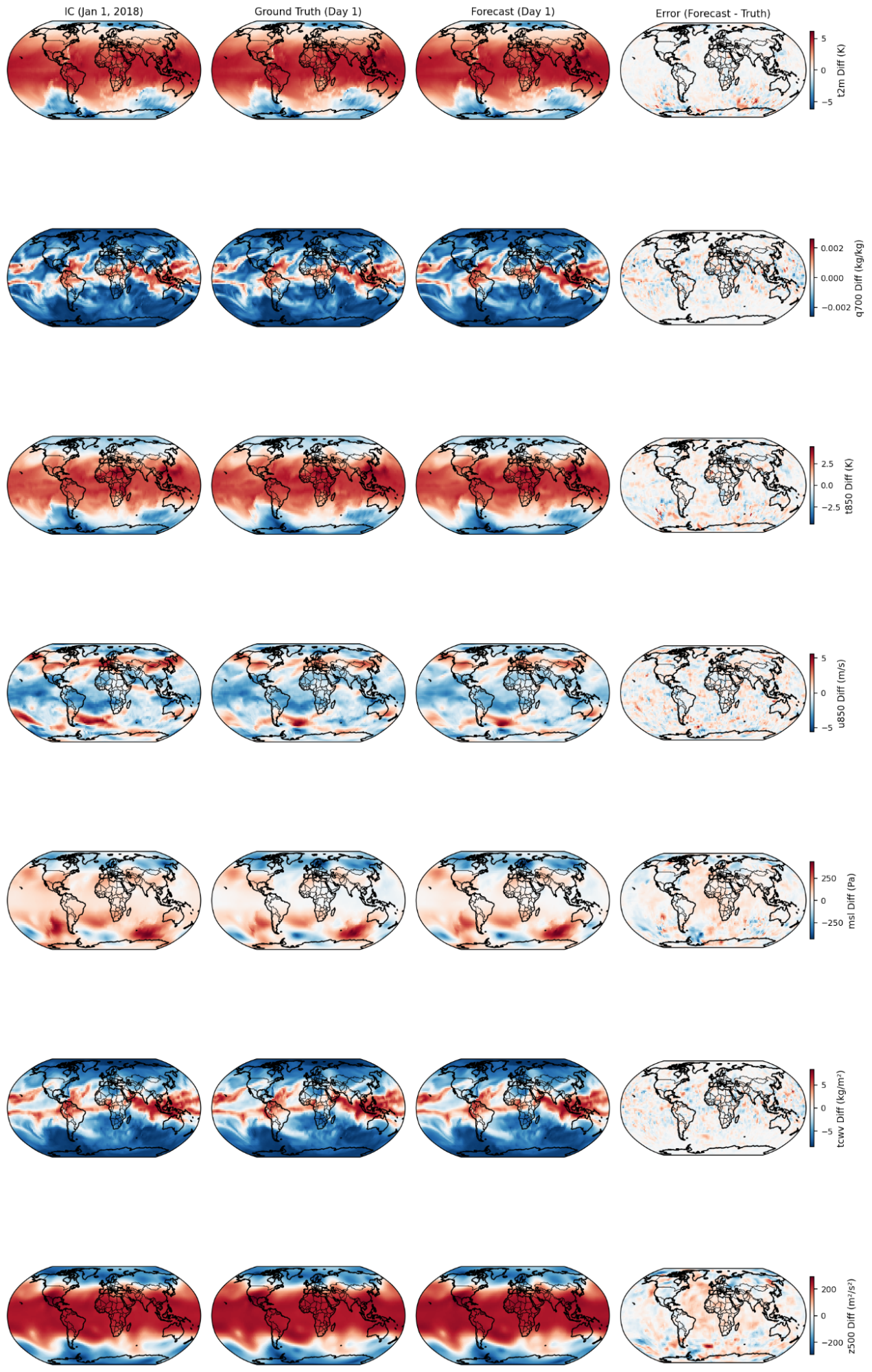}
	\caption{Forecast visualization at 1-day lead time. Each row shows a different atmospheric variable. Columns represent (1) initial condition (Jan 1, 2018), (2) ground truth, (3) forecast, and (4) forecast error (forecast - truth).}
	\label{fig:kai_day1}
\end{figure}
\FloatBarrier

\begin{figure}[htbp]
	\centering
	\includegraphics[height=0.85\textheight]{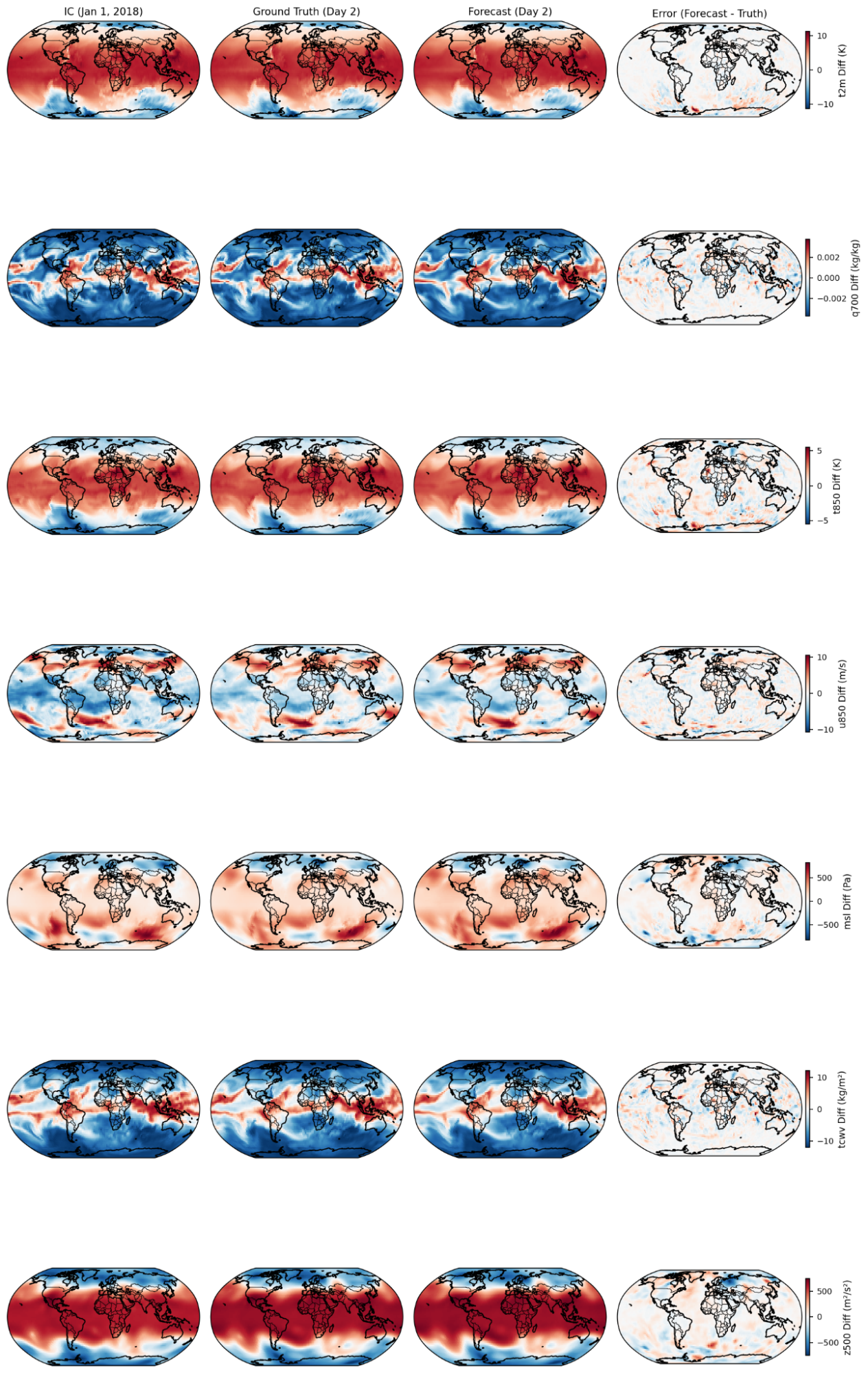}
	\caption{Forecast visualization at 2-day lead time. Each row shows a different atmospheric variable. Columns represent initial condition, ground truth, forecast, and error.}
	\label{fig:kai_day2}
\end{figure}
\FloatBarrier

\begin{figure}[htbp]
	\centering
	\includegraphics[height=0.85\textheight]{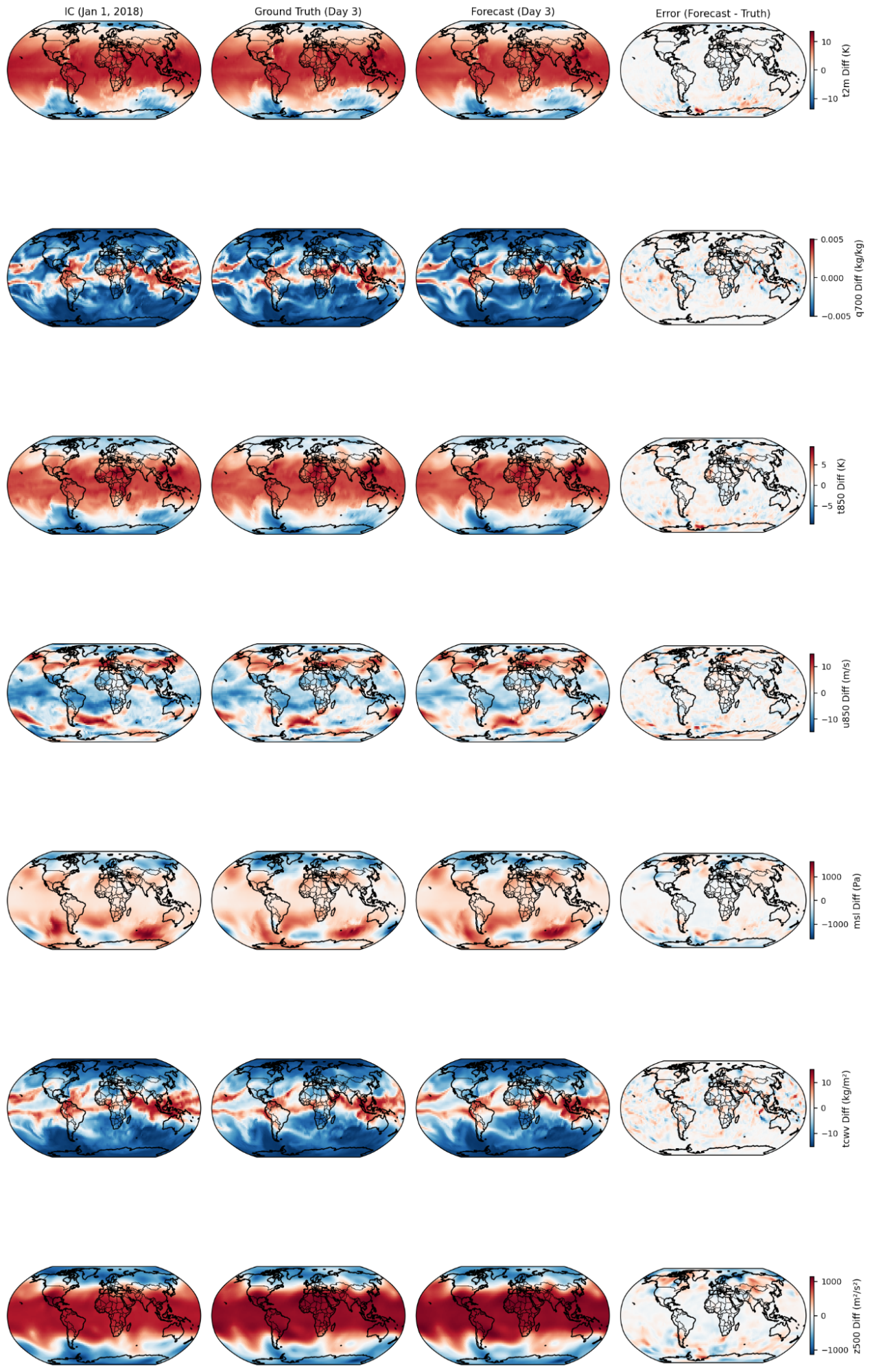}
	\caption{Forecast visualization at 3-day lead time. Each row shows a different atmospheric variable. Columns represent initial condition, ground truth, forecast, and error.}
	\label{fig:kai_day3}
\end{figure}
\FloatBarrier

\begin{figure}[htbp]
	\centering
	\includegraphics[height=0.85\textheight]{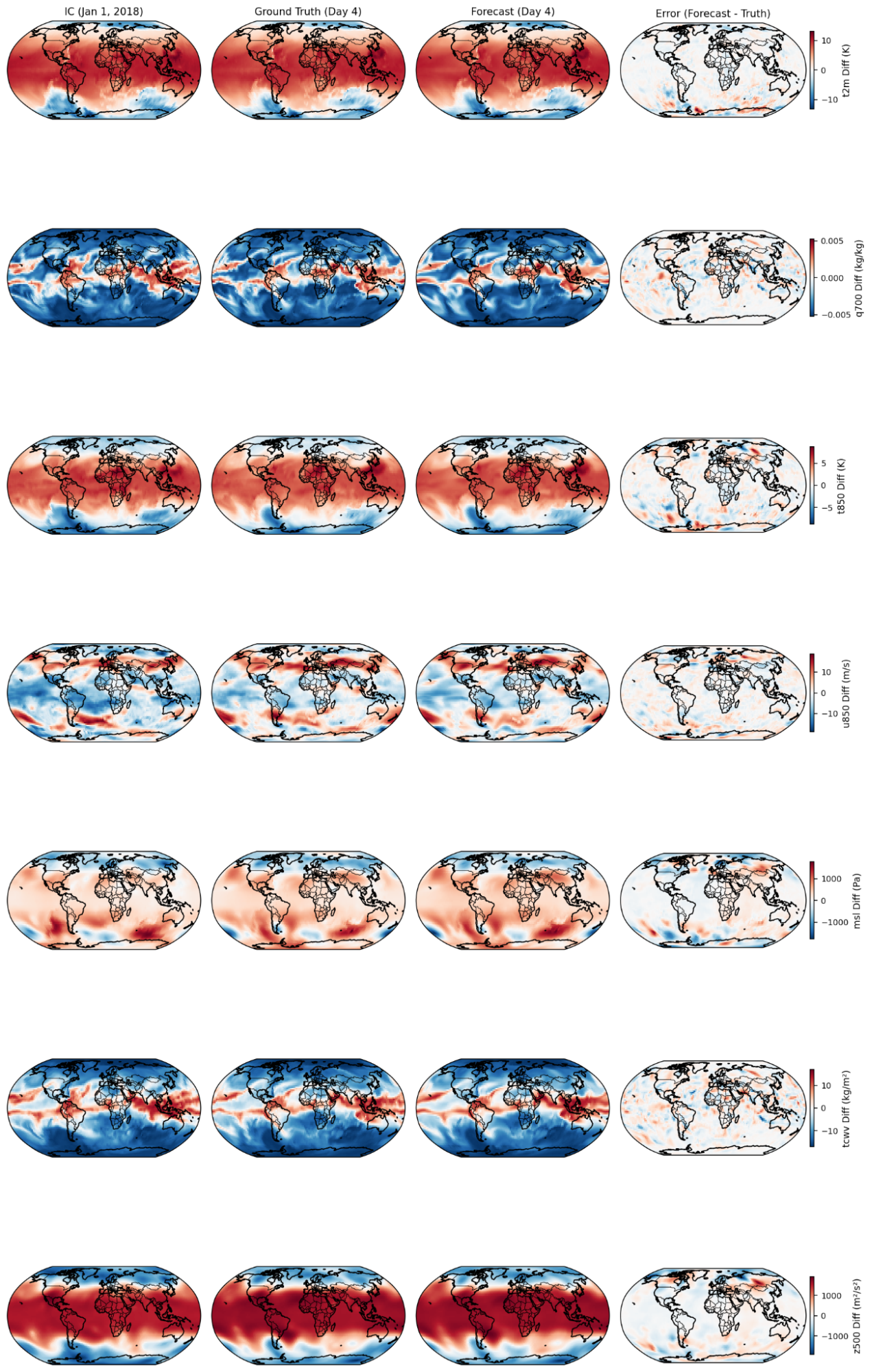}
	\caption{Forecast visualization at 4-day lead time. Each row shows a different atmospheric variable. Columns represent initial condition, ground truth, forecast, and error.}
	\label{fig:kai_day4}
\end{figure}
\FloatBarrier

\begin{figure}[htbp]
	\centering
	\includegraphics[height=0.85\textheight]{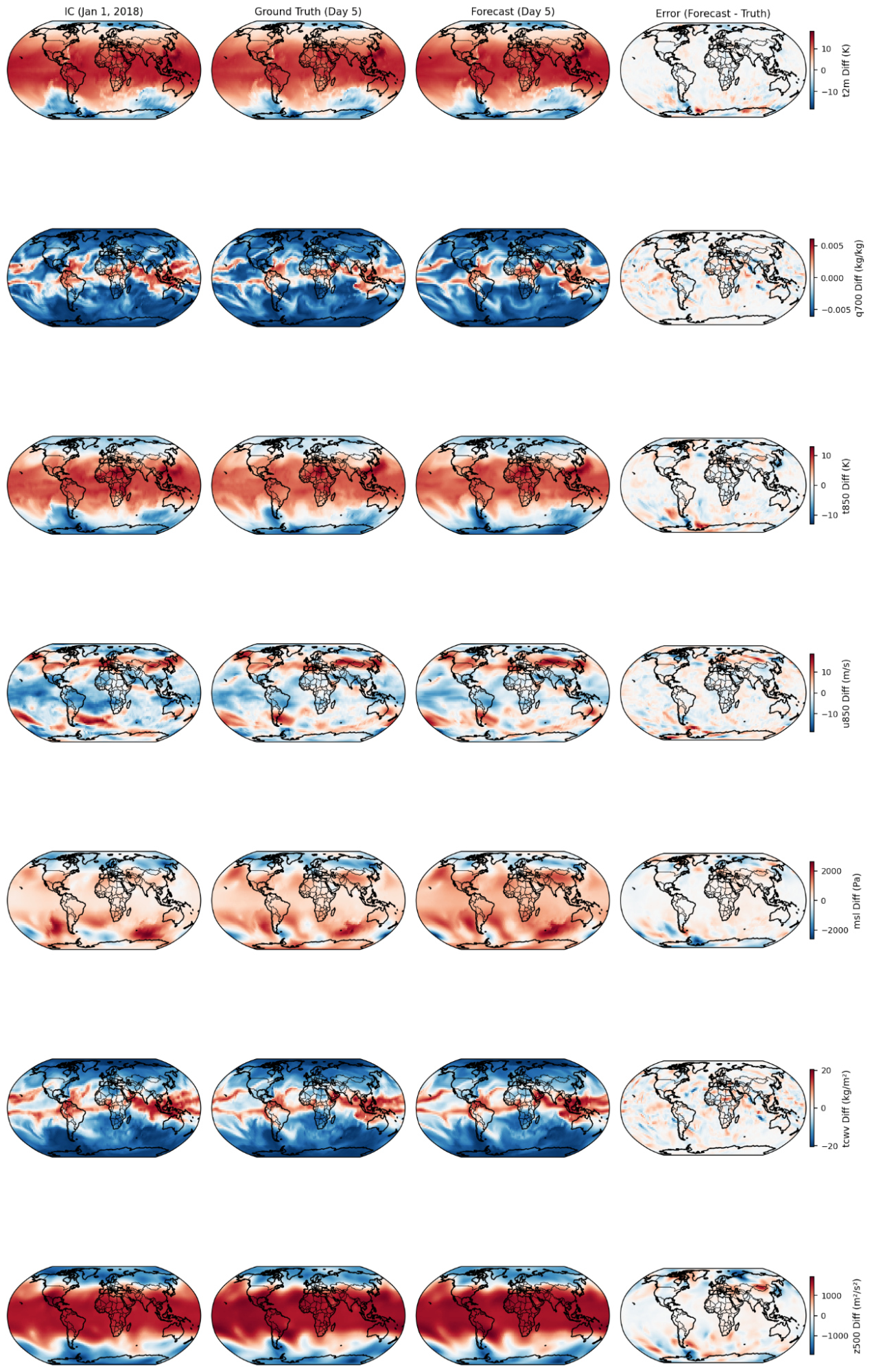}
	\caption{Forecast visualization at 5-day lead time. Each row shows a different atmospheric variable. Columns represent initial condition, ground truth, forecast, and error.}
	\label{fig:kai_day5}
\end{figure}
\FloatBarrier

\begin{figure}[htbp]
	\centering
	\includegraphics[height=0.85\textheight]{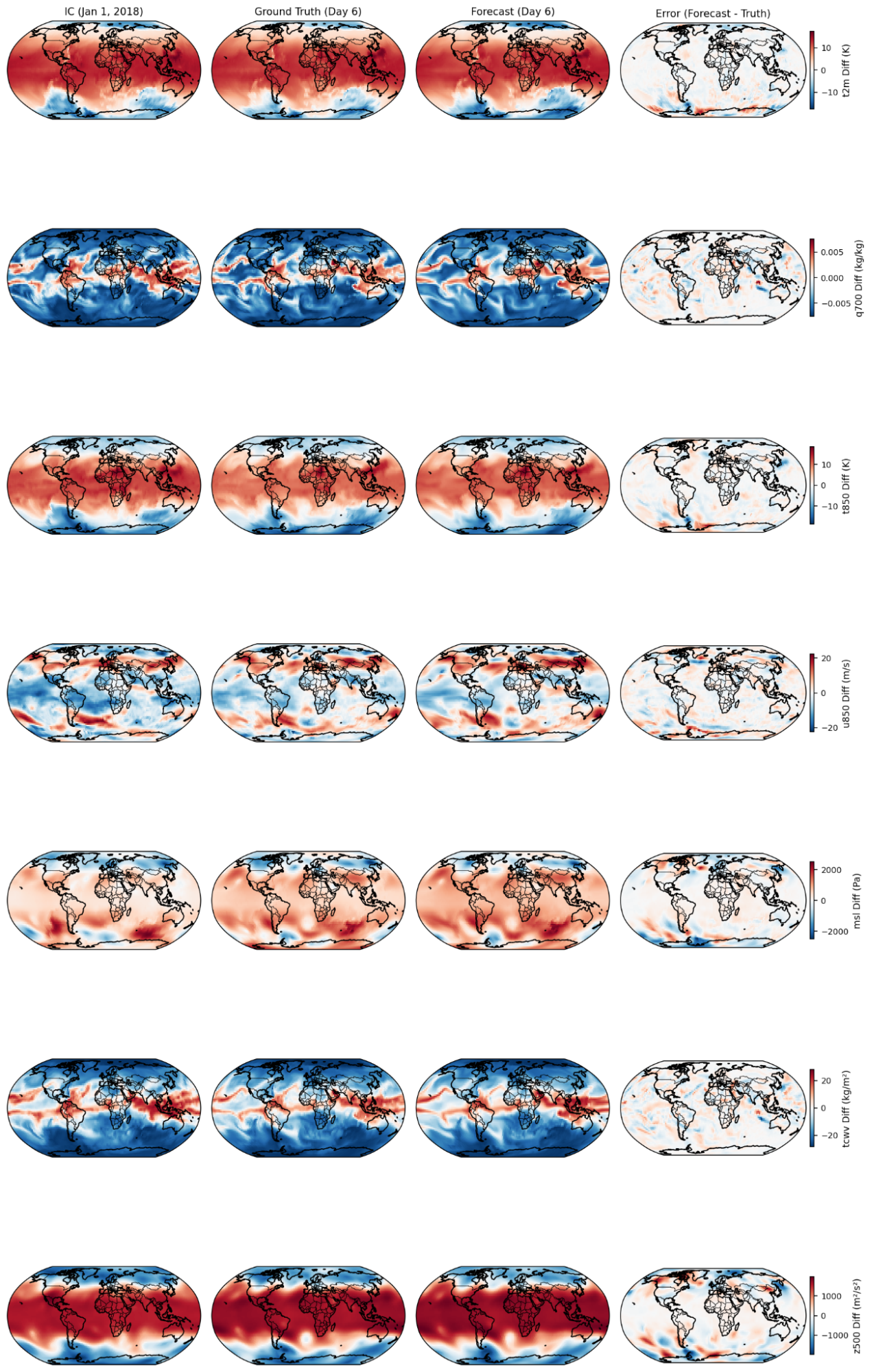}
	\caption{Forecast visualization at 6-day lead time. Each row shows a different atmospheric variable. Columns represent initial condition, ground truth, forecast, and error.}
	\label{fig:kai_day6}
\end{figure}
\FloatBarrier

\begin{figure}[htbp]
	\centering
	\includegraphics[height=0.85\textheight]{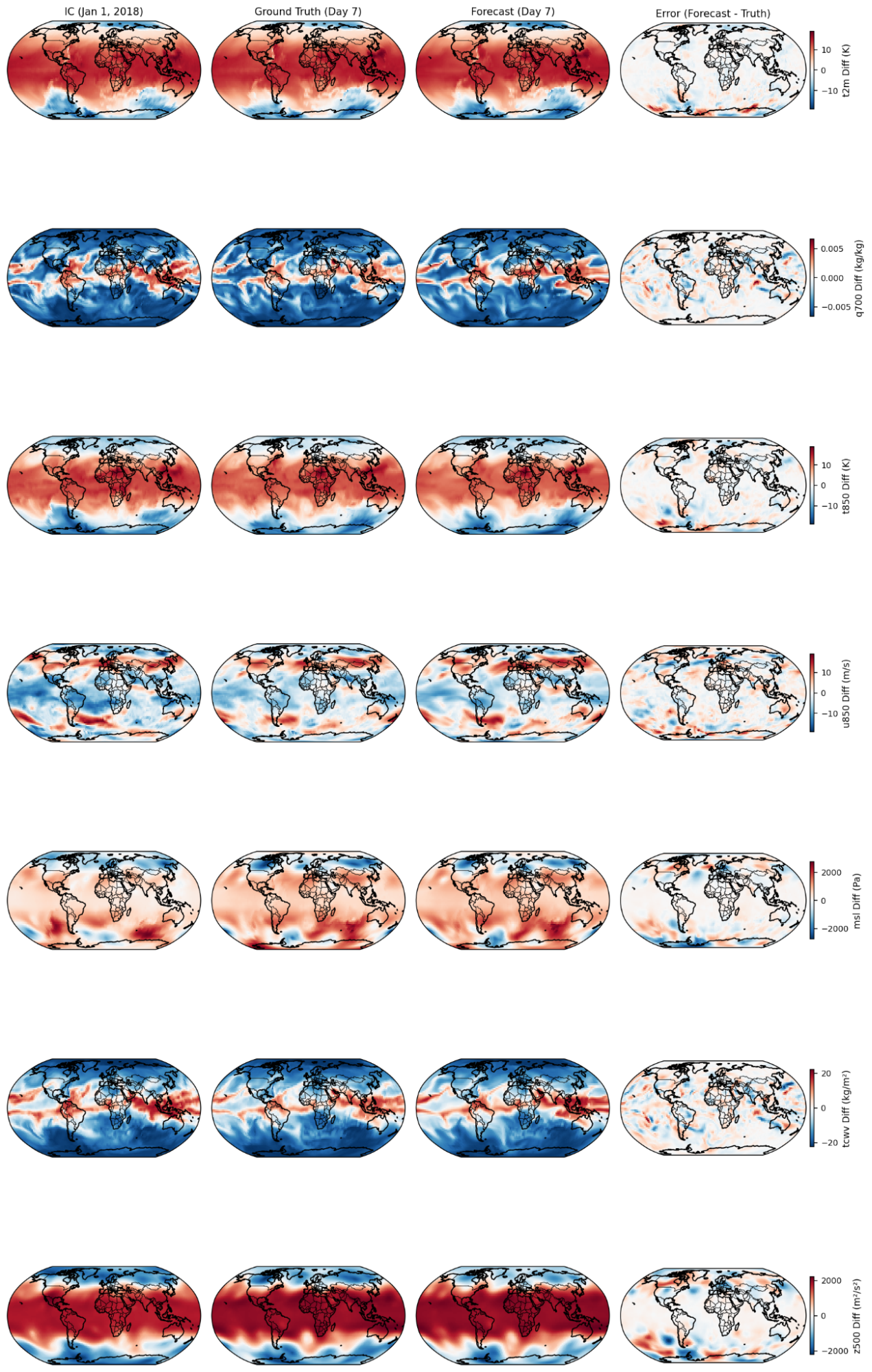}
	\caption{Forecast visualization at 7-day lead time. Each row shows a different atmospheric variable. Columns represent initial condition, ground truth, forecast, and error.}
	\label{fig:kai_day7}
\end{figure}
\FloatBarrier

\begin{figure}[htbp]
	\centering
	\includegraphics[height=0.85\textheight]{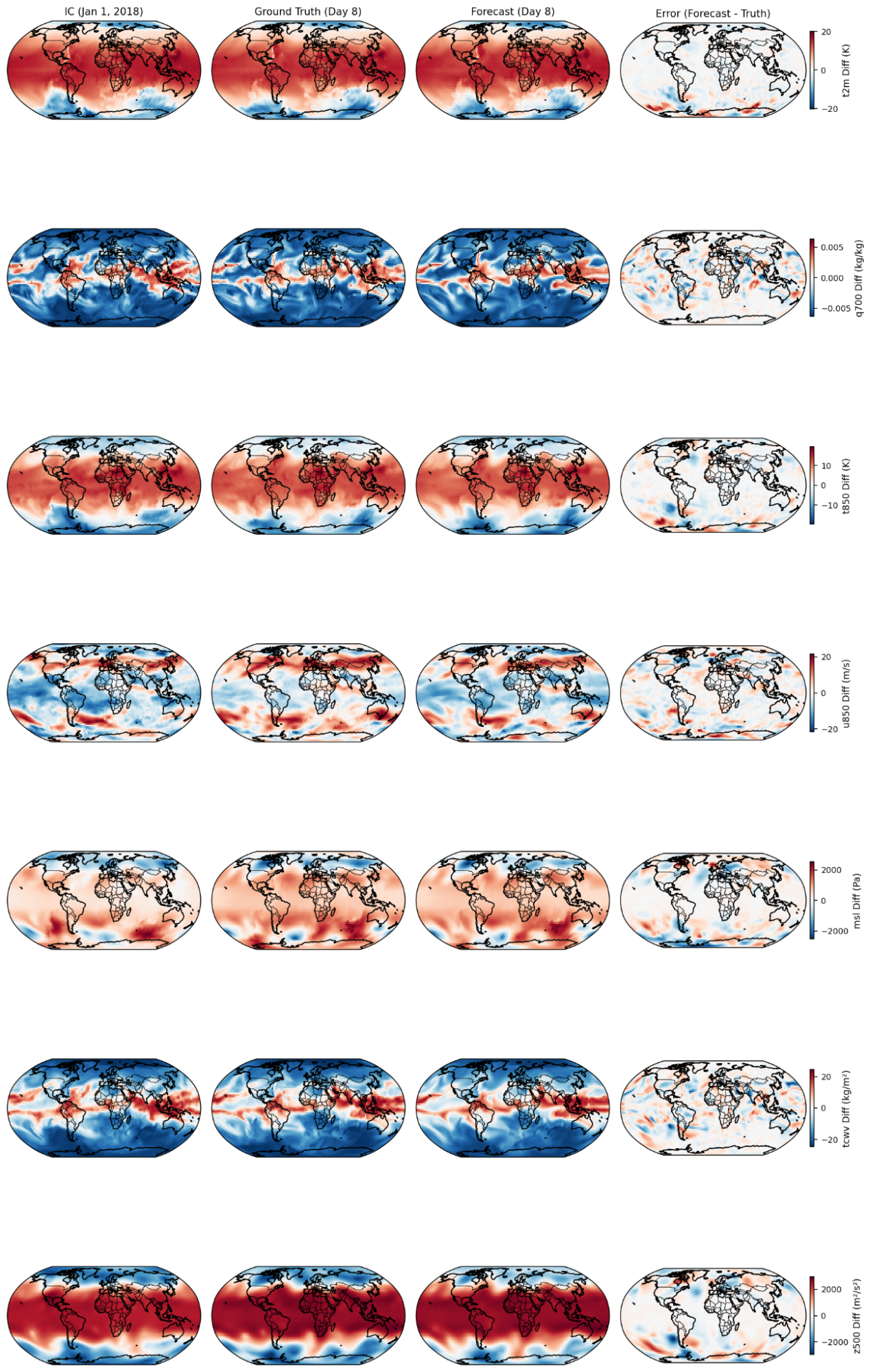}
	\caption{Forecast visualization at 8-day lead time. Each row shows a different atmospheric variable. Columns represent initial condition, ground truth, forecast, and error.}
	\label{fig:kai_day8}
\end{figure}
\FloatBarrier

\begin{figure}[htbp]
	\centering
	\includegraphics[height=0.85\textheight]{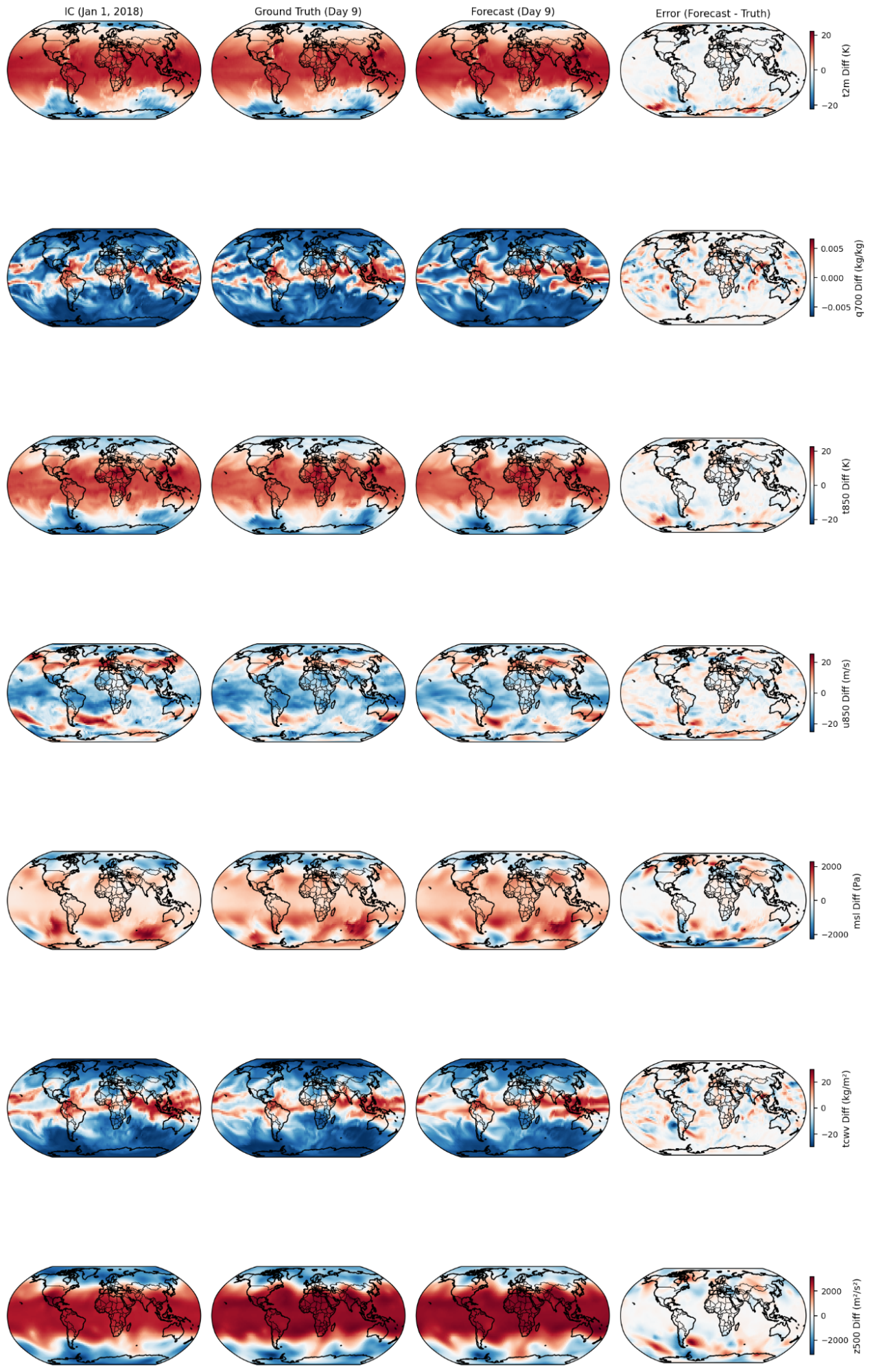}
	\caption{Forecast visualization at 9-day lead time. Each row shows a different atmospheric variable. Columns represent initial condition, ground truth, forecast, and error.}
	\label{fig:kai_day9}
\end{figure}
\FloatBarrier

\begin{figure}[htbp]
	\centering
	\includegraphics[height=0.85\textheight]{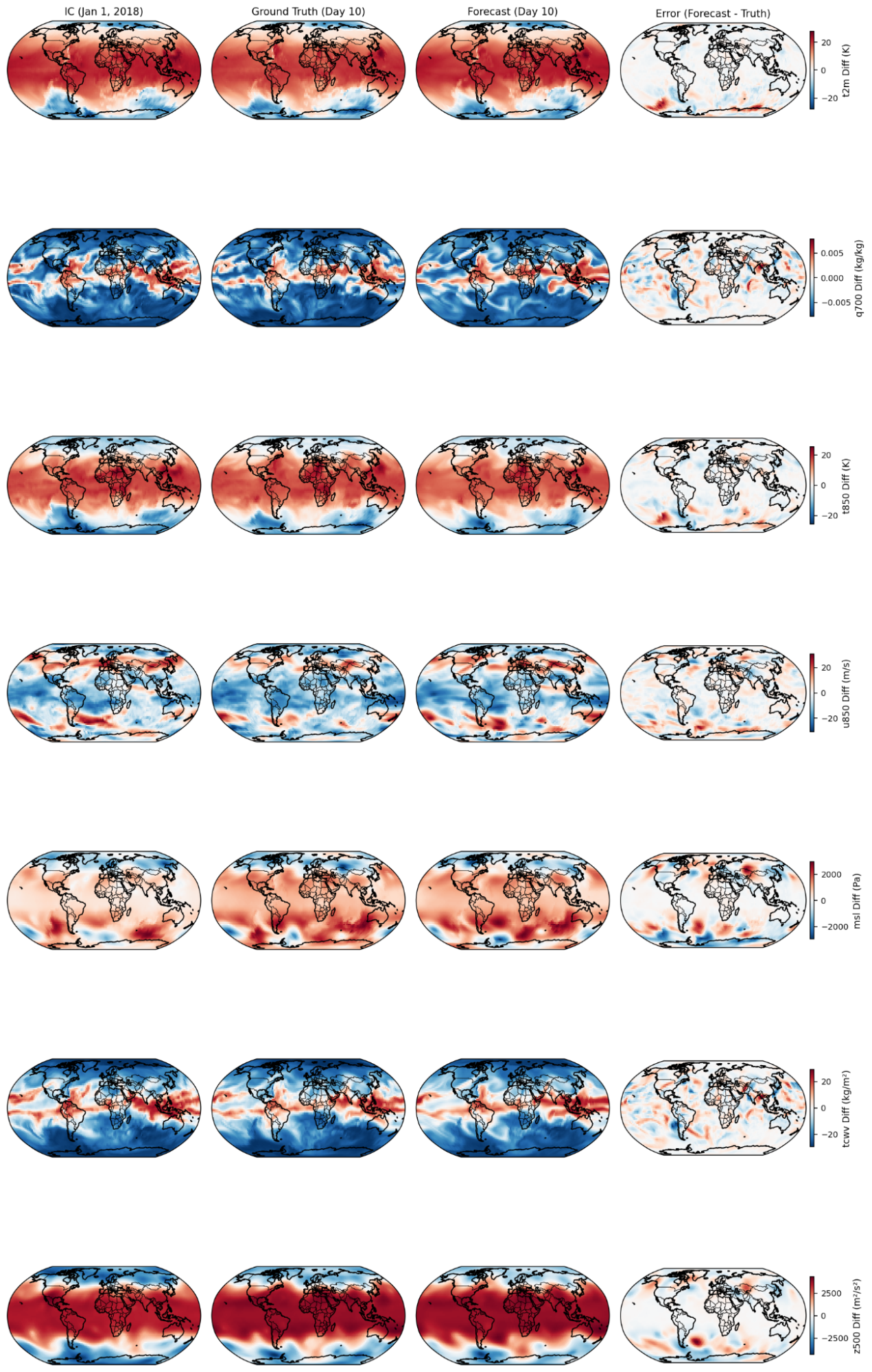}
	\caption{Forecast visualization at 10-day lead time. Each row shows a different atmospheric variable. Columns represent initial condition, ground truth, forecast, and error.}
	\label{fig:kai_day10}
\end{figure}
\FloatBarrier

\begin{table}[h!]
\centering
\caption{Number of parameters in recent AI-based weather forecasting models.}
\label{tab:model_params_alpha}
\begin{tabular}{@{}lc@{}}
\toprule
\textbf{Model} & \textbf{Parameters} \\
\midrule
AIFS-CRPS             & 229M \\
ArchesWeather-M       & 84M \\
ArchesWeather-S       & 44M \\
Aurora                & 1.3B \\
ClimaX                & 207M \\
ClimODE               & 2.8M \\
ExtremeCast           & 570M \\
FengWu-GHR            & 4B \\
GraphCast             & 36.7M \\
MetMamba              & 20M \\
MetNet                & 225M \\
MetNet-3              & 227M \\
Pangu-Weather         & 256M \\
Prithvi               & 100M \\
Prithvi WxC           & 2.3B \\
SEEDS-GEE             & 114M \\
\midrule
\textbf{KAI-$\alpha$} & \textbf{7M} \\
\bottomrule
\end{tabular}
\end{table}

\end{document}